\documentclass[10pt,twocolumn,letterpaper]{article}

\usepackage{cvpr}              

\usepackage{graphicx}
\usepackage{amsmath}
\newtheorem{myDef}{Definition}
\usepackage{autobreak}
\usepackage{amssymb}
\usepackage{booktabs}
\usepackage[pagebackref,breaklinks,colorlinks]{hyperref}

\usepackage{times}
\usepackage{epsfig}
\usepackage{ulem}
\usepackage{mathrsfs}
\usepackage[table,xcdraw]{xcolor}
\usepackage{color}
\usepackage{url}
\usepackage{siunitx}
\usepackage{booktabs}
\usepackage{tabularx}
\usepackage{float}
\usepackage{algorithm}
\usepackage{algpseudocode}
\usepackage{multirow}

\usepackage[capitalize]{cleveref}
\crefname{section}{Sec.}{Secs.}
\Crefname{section}{Section}{Sections}
\Crefname{table}{Table}{Tables}
\crefname{table}{Tab.}{Tabs.}

\begin{document}

\title{Learning from Pixel-Level Noisy Label : A New Perspective for \\
	Light Field Saliency Detection}
\author{Mingtao Feng$^{1}$\thanks{Equal contribution}~~ Kendong Liu$^{1 *}$~~ Liang Zhang$^{1 }$\thanks{Corresponding author}~~ Hongshan Yu$^{2}$~~ Yaonan Wang$^{2}$~~ Ajmal Mian$^{3}$\\
		$^1$Xidian University,~~ $^2$Hunan University,~~ $^3$The University of Western Australia\\
	}

\maketitle

\begin{abstract}

\vspace{-3mm}

Saliency detection with light field images is becoming attractive given the abundant cues available, however, this comes at the expense of large-scale pixel level annotated data which is expensive to generate.
In this paper, we propose to learn light field saliency from pixel-level noisy labels obtained from unsupervised hand crafted featured-based saliency methods. Given this goal, a natural question is: can we efficiently incorporate the relationships among light field cues while identifying clean labels in a unified framework? We address this question by formulating the learning as a joint optimization of intra light field features fusion stream and inter scenes correlation stream to generate the predictions. Specially, we first introduce a pixel forgetting guided fusion module to mutually enhance the light field features and exploit pixel consistency across iterations to identify noisy pixels. Next, we introduce a cross scene noise penalty loss for better reflecting latent structures of training data and enabling the learning to be invariant to noise. Extensive experiments on multiple benchmark datasets demonstrate the superiority of our framework showing that it learns saliency prediction comparable to state-of-the-art fully 
supervised light field saliency methods. Our code is available at~\url{https://github.com/OLobbCode/NoiseLF.} 

\end{abstract}
\vspace{-6mm}

\section{Introduction}
\vspace{-2mm}

\begin{figure}[t]
	\centering
	\includegraphics[width=\linewidth]{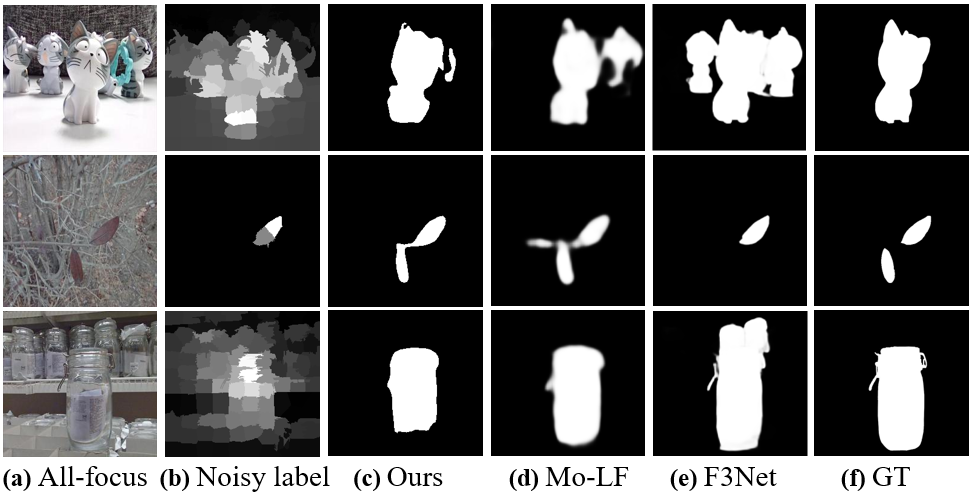}
	\vspace{-6mm}
	\caption{Saliency detection in challenging light field scenes. (a) all-focus images; (b) pixel-level noisy labels; (c)-(e) results of our method, fully supervised light field method Mo-LF~\cite{Memory_Nips} and RGB method F3Net~\cite{F3Net}; (f) ground truth saliency maps, depicted only for illustration purposes and not used in our training.}
	\label{fig_intro}
	\vspace{-6mm}
\end{figure}

Saliency detection imitates the human attention mechanism and allows us to focus on the most visually distinctive regions out of an overwhelming amount of information. This problem has attracted much attention given the broad applications in computer vision, such as image and video segmentation, visual tracking and robot navigation~\cite{SOD_review1,SOD_review2,SOD_review3}. Existing saliency detection methods can be roughly divided into three categories based on the 2D (RGB), 3D (RGB-D) and 4D (light field) input images. Unlike the former two, light field provides multi-view images of the scene through an array of lenslets and produces a stack of focal slices, containing abundant spatial parallax information as well as depth information~\cite{LF_1,LFSOD_CVPR21,LF_review}. Moreover, light field data consists of an all-focus central view and a focal stack, where the stack of focal slices (similar to human visual perception) are observed in sequence with a combination of eye movements and shifts in visual attention~\cite{Exploit_LF_fusion}. Such a comprehensive 4D data provides abundant cues for saliency detection in challenging scenes e.g. similar foreground and background, small salient objects and complex background, as shown in Figure~\ref{fig_intro} (a).

Early light field saliency detection works have been dominated by fully supervised methods which require large amounts of accurate pixel-level annotations aligned with the all-focus central view for training~\cite{Deep_LF_fusion,Memory_Nips,LFSOD_LFNet,LFSOD_cellular,LFSOD_muti-task}. This expensive and time-consuming labelling process hinders the applicability of fully supervised methods to large scale problems. If the tedious pixel-level annotation process can somehow be avoided, we can exploit the unlimited supply of light field images from hand-held cameras (such as Lytro Illum~\cite{Lytro} and Raytrix~\cite{Raytrix}) for large scale applications. In this paper, we are interested in learning light field saliency prediction from single per-pixel noisy labels, where the per-pixel noisy labels are produced by existing low cost off-the-shelf conventional unsupervised saliency detection methods. These labels are noisy compared to the ground truth human annotations and can have method-specific bias in predicting the saliency map. In our configuration, for each light field image in the training data, only a single noisy saliency map is available. 

Directly training the light field saliency detection network on the pixel-level noisy labels may guide the network to overfit to the corrupted labels~\cite{RGB_salience_PAMI}. Additionally, previous light field saliency detection methods lack a global perspective to explore patterns in the relationships between the whole dataset. To effectively leverage these noisy but informative saliency maps, we propose a new perspective to the light field saliency detection problem: \textit{how to efficiently incorporate the relations among light field cues while identifying clean labels in a unified framework?} To this end, we make two major contributions, intra light field features fusion and across scenes correlation. \textit{Firstly}, we introduce a pixel forgetting guided fusion module to explore the interactions among all-focus central view image and focal slices, and exploit pixel consistency across training iterations to identify noisy pixels. Specially, we perform the interaction process in a reciprocating fashion, where mutual guidance first emphasizes the useful features and suppresses the unnecessary ones from focal slices using the all-focus central view. Next, the weighted focal stack features are used to gradually refine the spatial information of all-focus central view for accurately identifying salient objects. For the initial noisy estimations of the updated focal stack features and all-focus central view features, we introduce pixel forgetting event to evolve across the training iterations and define a forgetting matrix to identify noisy pixels. The final prediction comprises pixels with high certainty from the initial noisy estimations. Thus we can simultaneously explore the abundant light field cues and identify inliers for our model. 

\textit{Secondly}, we propose a cross-scene noise penalty loss to capture the global correlation of the noise space for better reflecting intrinsic structures of the whole training data to enable more robust saliency predictions. The first term of our cross-scene noise penalty loss evaluates the network's prediction on training light field images using noisy labels, and the second term is defined on several independent randomly selected light field images to penalize the networks from overly agreeing with the pixel-level noisy labels. Both terms encode the knowledge of noise rates implicitly and allow our light field  saliency prediction model to become invariant to pixel-level noise.

To the best of our knowledge, this is the first work that proposes the idea of considering light field saliency detection as learning from pixel-level noisy labels which is a completely different direction from existing fully supervised methods. Our main contributions are: (1) We formulate the saliency prediction as a joint optimization of intra light field features fusion stream and inter scenes correlation stream. (2) We introduce a pixel forgetting guided fusion module to mutually enhance the light field features and exploit pixel consistency across iterations to identify noisy pixel labels. (3) We propose a cross-scene noise penalty loss for better reflecting latent structures of training data and enabling the learning to be invariant to label noises. We perform thorough experimental evaluations of the proposed model, which achieves comparable performance with state-of-the-art fully supervised light field saliency prediction methods.

\vspace{-2mm}
\section{Related works}
\vspace{-2mm}

\noindent \textbf{Light field saliency detection:} Conventional methods for light field salient object detection often extend various hand-crafted features (e.g., global/local color contrast, background priors and object location cues) and adapt tailored light field features (e.g., focusness and depth) to the case of light field data~\cite{LF_review}. Li et al.~\cite{T_LFSaliency_2014} proposed the pioneering and earliest work on light field saliency detection, which incorporates the focusness measure with location priors to determine the background and foreground slices. DILF (Deeper Investigation of Light Field)~\cite{T_LFSaliency_Ijcai15} computes the background priors based on the focusness measure embedded in the focal stacks and uses them as weights to eliminate background distraction and enhance saliency estimation. Piao et al.~\cite{T_LFSaliency_TIP} introduced a depth-induced cellular automata for light field saliency object detection, and then a Bayesian fusion strategy and CRF~\cite{PAMI_CRF} are employed to refine the prediction. 

Due to their powerful learning ability, several deep learning methods have promoted the light field saliency detection performance significantly. Zhang et al.~\cite{Memory_Nips} proposed a memory-oriented spatial fusion module to go through all pieces of focal stack and all-focus features. However, their method only fuses focal stack and all-focus features once. Wang et al.~\cite{Deep_LF_fusion} and Piao et al.~\cite{Exploit_LF_fusion} both fused the features from different focal slices using varying attention weights, which are inferred at multiple time steps in a ConvLSTM. As such, they performed feature fusion within focal slices several times. However, \cite{Deep_LF_fusion} conducted focal stack and all-focus features fusion only once, while \cite{Exploit_LF_fusion} did not perform such a fusion~\cite{LFSOD_CVPR21}. They adopted knowledge distillation to improve the representation ability of the all-focus branch. 

In contrast to existing methods, we propose a robust technique to learn from noisy annotated light field data for saliency detection. To the best of our knowledge, we are the first to formulate pixel-level noisy label learning as a joint optimization of intra light field image fusion and inter light field image correlation streams to generate the predictions.

\noindent \textbf{Learning From Noisy Labels:} A large portion of research on learning with noisy labels is dedicated to learning classification models in the presence of inaccurate class labels. To handle noisy labels, three main directions have been explored: 1) developing regularization techniques~\cite{Noisy_1}; 2) estimating the noise distribution~\cite{Noisy_2,Noisy_3}; 3) training on selected samples~\cite{Noisy_4,Noisy_5,Noisy_6,Noisy_7}. All these methods deal with image classification. Existing works in learning from dense noisy labels require multiple noisy versions of pixel-wise labelling for each input image~\cite{RGB_salience_fusion,RGB_salience_multi,RGB_salience_nips}.
Zhang et al.~\cite{RGB_salience_fusion} fused saliency maps from unsupervised handcrafted feature-based methods with heuristics within a deep learning framework. The recursive optimization in~\cite{RGB_salience_multi} depends on a two-stage mechanism for refinement of pseudo-labels and saliency detection network.

Nguyen et al.~\cite{RGB_salience_nips} defined image-level loss function to train with noisy labels for generating a coarse saliency map, and then iteratively refined it with moving average and fully connected CRF. Unlike \cite{RGB_salience_fusion,RGB_salience_multi,RGB_salience_nips}, \cite{RGB_salience_PAMI,RGB_salience_Eccv} deals with learning from a single noisy labelling in a much more efficient way. \cite{RGB_salience_PAMI} learns saliency prediction and robust fitting models to identify inliers. \cite{RGB_salience_Eccv} proposes to learn a clean saliency predictor from a single noisy label by latent variable model called noise-aware encoder-decoder. We take a completely new approach and propose a principled method for dealing with the dense prediction task of saliency detection in challenging 4D light field scenes. Our method efficiently incorporates the relationships among light field cues to correct pixel-level noisy labels.

\begin{figure*}[t]
	\centering
	\includegraphics[width=0.95\linewidth]{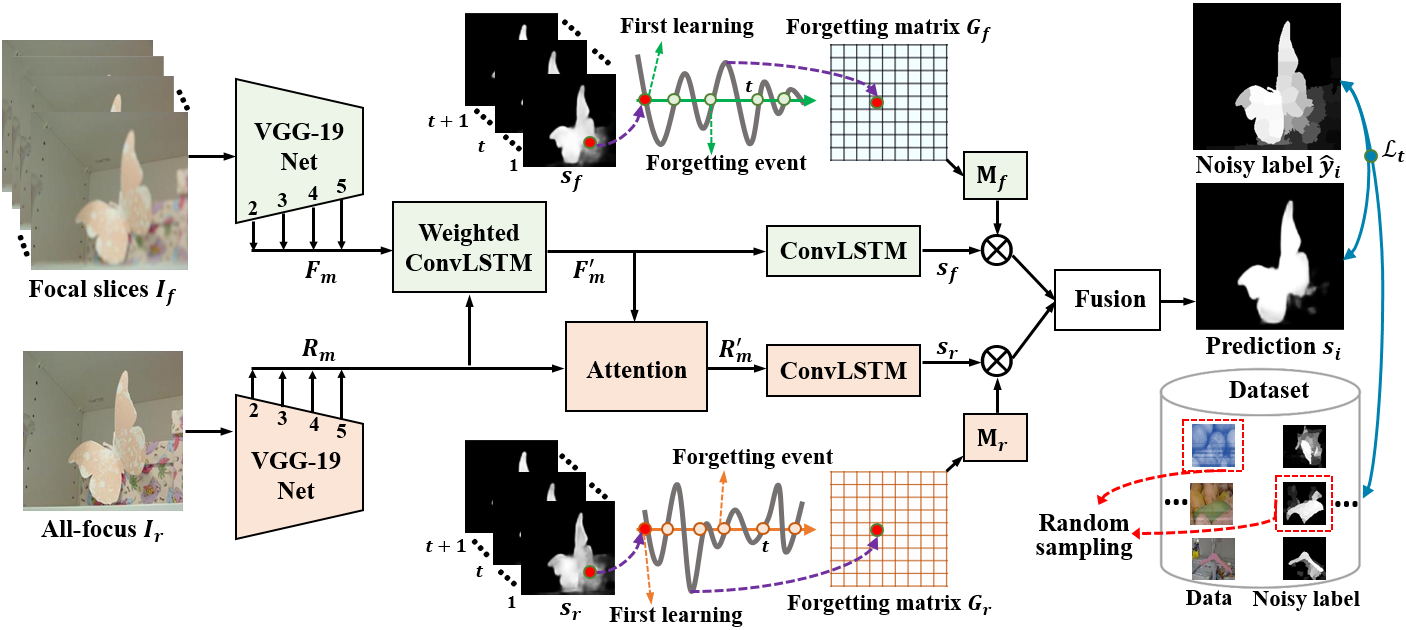}
	\vspace{-2mm}
	\caption{Overview of the complete architecture of our proposed network, which formulates the learning as a joint optimization of intra light field features fusion stream and inter scenes correlation stream to generate the prediction. }
	\label{fig_overview}
	\vspace{-2mm}
\end{figure*}


\vspace{-3mm}
\section{Proposed Approach}
\vspace{-2mm}
In this paper, we focus on learning light field saliency from a single pixel-level noisy map. 
Specifically, we aim to learn an accurate saliency map of a light field image $x_i$ given its pixel-level noisy saliency map $\hat{y}_i$ produced by an off-the-shelf low cost handcrafted features based method. 
A trivial and direct solution would be to use the noisy saliency map as `proxy' human annotations for training a deep model.
However, such an approach does not suffice since network learning is highly prone to noise in the supervision labels~\cite{Noise_network}. We propose a unified framework to incorporate the relationships among light field cues while identifying clean labels during training in a unified framework. Our complete pipeline is shown in Fig.~\ref{fig_overview} and the technical details of each component are elaborated below.

We start with a training set $\mathcal{T}=\{(x_i, \hat{y}_i)\}_{i=1}^N$, where each $x_i$ is a light field image with spatial size $u\times v$ and $\hat{y}_i\in [0,1]^{\mathrm{u\times v}}$ is its noisy binary saliency map. We use $\hat{y}_i$ (instead of $y_i$) to differentiate is from clean labels as in the human annotated labelling setting. 
For each $x_i$, we have an all-focus central view image $I_r$ and its corresponding focal stack $I_f$ with $k$ focal slices $\{I_{f_1},I_{f_2},\dots I_{f_k}\}$, which have different focused regions. A deep model for light field saliency learns a mapping function $f_{\Theta}: \{I_r, I_f\}\to [0,1]^{\mathrm{u\times v}}$, where $\Theta$ is a set of network parameters. Therefore, $f_{\Theta}(x_i)=s_i$ denotes the predicted saliency map.


\subsection{Pixel Forgetting Guided Fusion Module}

\noindent \textbf{Mutual feature fusion:} We adopt VGG-19~\cite{VGG} as the backbone architecture to generate all-focus central view image features $R_m$ and light field focal stack features $F_m=\{f_m^i\}_{i=1}^{k}$ with abundant spacial information~\cite{Memory_Nips}, where $m=2,3,4,5$ represent index of the high-level features from the last 4 convolution blocks of VGG-19. Instead of processing the two types of features separately, we build a mutual fusion strategy between the all-focus central view image features and the focal stack features. In each step, the former is first used to guide the update of the latter, and then the refined feature is used to update the former.

We start from fusing the focal stack features $F_m$ by propagating contexts within the focal slices and also under the guidance from the all-focus features $R_m$, which provides external guidance for the feature update of $F_m$. We use an attention mechanism to emphasize the useful features and suppress the unnecessary ones from focused and blurred information. This procedure can be defined as:
\vspace{-2mm}
\begin{equation}
{\rm Att}_m=\sigma( w_m\ast {\rm Avg}(C[R_m;f_m^1,f_m^2,\cdots, f_m^k])+b_m),
\end{equation}
\begin{equation}
\bar{f_m}^i=f_m^i\odot {\rm Att}_m^i,
\end{equation}
where $C[\cdot]$ means concatenation operation, and $i=1,2,\cdots, k$. $\ast, w_m$ and $b_m$ represent the convolution operator and convolution parameters in $m$-th layers. Avg($\cdot$) means global average pooling operation and $\sigma(\cdot)$ means softmax function. ${\rm Att}_m\in \mathbb{R}^{1\times1\times{(k+1)}}$ means the channel-wise attention map in the $m$-th layer and $\odot$ denotes feature-wise multiplication.

The weighted light field features $\bar{F}_m=\{\bar{f_m}^i\}_{i=1}^k$ are regarded as a sequence of inputs corresponding to consecutive time steps. They are fed into a ConvLSTM~\cite{ConvLSTM}
structure to gradually refine their spatial information for accurately identifying the salient objects. The refined focal slices features are represented by $F_m'$.

To allow the focal slices features $F_m'$ to guide the updating of all-focus central view image features $R_m$, we employ an attention mechanism to emphasize or suppress each pixel in $R_m$. 
\vspace{-3mm}
\begin{equation}
{\rm Att}_{F'_{m}}=\sigma(w\ast F_m'+b)
\end{equation}
\begin{equation}
R_m'=R_m\otimes{\rm Att}_{F'_{m}}+R_m
\end{equation}
where $\sigma(\cdot)$ is softmax function, $\ast, w$ and $b$ represent convolution operator and convolution parameters in $m$-th layers. ${\rm Att}_{F'_{m}}\in \mathbb{R}^{\mathrm{u\times v}}$ means the pixel-wise attention map in $m$-th layer and  $\otimes$ denotes pixel-wise multiplication.

Furthermore, efficient integration of all the $m$ hierarchical updated features can improve discriminability and is significant for the saliency prediction task. The updated feature $F_m'$ and $R_m'$ are separately fed into ConvLSTM cells to further summarize the spatial information. The output of the ConvLSTM is followed by a transition convolutional layer and an up-sampling operation to get the initial noisy predictions $s_f$ and $s_r$ for the focal slices and the all-focus central view image branch respectively.

\noindent \textbf{Pixel consistency:} We gain insights into the optimization process by analyzing dynamic pixels of the initial noisy predictions $s_f$ and $s_r$ during learning and their influence on the final prediction in the saliency map. We observe that the noisy pixels exhibit characteristics different from clean pixels. We would expect that noisy-labeled pixels will learn constantly due to the inconsistency with dominant decision across iterations, which is validated later in experiments. These atypical characteristics of noisy pixels support the identification of clean pixels in the initial noisy predictions $s_f$ and $s_r$. Inspired by~\cite{Forgetting}, we further define forgetting events to have occurred for pixels of $s_f$ or $s_r$ when 
they 
transition from being recognized correctly (salient object) to incorrectly (background) over the course of learning. These events occur when pixels that have been learnt at the $t$-th iteration are subsequently mistakenly recognized at the $(t+1)$-th iteration. We firstly define two transformation matrices $T_f$ and $T_r$ to describe the learning transform of pixels across the training phase, which exploit pixel consistency across iterations of initial noisy predictions $s_f$ and $s_r$.  
\begin{equation}
T_f(u,v)=
\begin{cases} 
0,&|s_f^{(u,v)}-\hat{y}^{(u,v)}|>\delta\\
1,&|s_f^{(u,v)}-\hat{y}^{(u,v)}|\leqslant\delta
\end{cases}
\end{equation}
\begin{equation}
T_r(u,v)=
\begin{cases} 
0,&|s_r^{(u,v)}-\hat{y}^{(u,v)}|>\delta\\
1,&|s_r^{(u,v)}-\hat{y}^{(u,v)}|\leqslant\delta
\end{cases}
\label{Euqation_delta}
\end{equation}
where $\hat{y}^{(u,v)}$ represents the noisy label of pixel $(u,v)$. The margin $\delta$ is defined as the bias between the logit of the initial noisy predictions and the noisy labels, which is further discussed in the experiments. These two transformation matrices are binary, indicating whether the pixels are correctly recognized at each epoch, and updated across iterations by a bias between the initial predictions and supervision information. 

Next, we introduce the forgetting matrix $G$ to compute forgetting event statistics for each pixel of the initial predicted noisy saliency maps $s_f$ and $s_r$. 
\begin{equation}
G_f(u,v)=
\begin{cases} 
G_f(u,v)+1,&T_f(u,v)^{t+1}<T_f(u,v)^{t}\\
G_f(u,v),&T_f(u,v)^{t+1}\geqslant T_f(u,v)^{t}
\end{cases}
\end{equation}
\begin{equation}
G_r(u,v)=
\begin{cases} 
G_r(u,v)+1,&T_r(u,v)^{t+1}<T_r(u,v)^{t}\\
G_r(u,v),&T_r(u,v)^{t+1}\geqslant T_r(u,v)^{t}
\end{cases}
\end{equation}
where we initialize $G$ as an all-zero matrix. Pixel $(u,v)$ in an image experiences a forgetting event when the corresponding element in the transform matrix decreases between two consecutive updates. In other words, pixel $(u,v)$ is recognized incorrectly at $(t+1)$-th iteration after having been correctly recognized at $(t)$-th iteration. When one pixel has frequent forgetting events during the training process, the value of its corresponding element in $G$ will increase cumulatively, indicating that the pixel is dynamic and noisy. 

Equipped with the forgetting matrix $G$, we further use a confidence re-weighting strategy to assign a soft weight to the initial predicted noisy saliency maps $s_f$ and $s_r$.
\begin{equation}
M_f(u,v)=\frac{2}{1+e^{a\cdot G_f^2(u,v)}}
\end{equation}
\begin{equation}
M_r(u,v)=\frac{2}{1+e^{a\cdot G_r^2(u,v)}}
\end{equation}
where $a$ is used to control the descent degree of the confidence weights according to the number of forgetting events. We set $a=0.04$ in our experiments. The soft weight matrix $M$ encourages pixels with consistent behavior to contribute more than those with dynamic behavior.

The final predicted saliency map $s_i$ can be obtained by the fusion of the initial predicted noisy saliency maps $s_f$ and $s_r$ under the guidance of pixel forgetting:
\begin{equation}
s_i= {\rm Up} (w\ast C[M_f\otimes s_f; M_r\otimes s_r]+b)
\end{equation}
where $C[\cdot]$ means concatenation operation, and $i=1,2,\cdots, N$. $\ast, w$ and $b$ represent convolution operator and convolution parameters. ${\rm Up}$ denotes the up-sample operation to get the final saliency map $s_i$.


\subsection{Cross-scene Noise Penalty Loss}
Previous light field saliency detection methods lack a global perspective to explore patterns in the relationships between the whole dataset. Modern approaches typically formulate saliency detection as a per-pixel classification task. Specially, $f_{\Theta}(x_i)=s_i$ denotes the predicted saliency map from the pixel forgetting guided fusion module, and the empirical risk when learning directly from noisy label can be defined as follows:
\begin{equation}
\mathcal{L}(s_i,\hat{y}_i)=\sum_{(u,v)}l(s_i^{(u,v)},\hat{y}_i^{(u,v)}),
\label{cross_entropy}
\end{equation}
where $X=\{x_i\}_{i=1}^N$, $\hat{Y}=\{\hat{y}_i\}_{i=1}^N$, and $(u,v)$ denotes the pixel spatial coordinates in a light field image, and $l: [0,1]\times [0,1]\to \mathbb{R}$ is the cross entropy loss defined as:
\begin{equation}
l(s,\hat{y})=-(\hat{y}\log(s)+(1-\hat{y})\log(1-s)),
\end{equation}

For a pixel wise prediction task, the network is trained by minimizing the per-pixel defined loss function. The optimal network model is obtained by minimizing Eq.~(\ref{cross_entropy}) using stochastic gradient descent. However, directly training the network with single noisy labelling will not work as it is well-known that network training is highly prone to noise in the supervision labels which may guide the network to overfit to the corrupted labels~\cite{Noise_network}.

Inspired from the correlation agreement (CA) mechanism~\cite{CA1}~\cite{CA2}, to guarantee the efficient learning of network training by  noisy labels, we elicit information from 
%
predictions performed over other samples of the data 
%
and score the current predictions against those. We seek to exploit correlations between current predictions with other scenes to align incentives with correct information, which we name as cross-scene evaluation in our method. 
\begin{myDef}
$\Delta\in\mathbb{R}^{2\times 2}$ is a square matrix with entries defined between the predicted saliency map $s_i$ and the noisy label $\hat{y}_i$, and characterizes the marginal correlations between them:
\begin{align}
\begin{autobreak}
\Delta_{a,b}=
p(s_i^{(u,v)}=J_{a},\hat{y}_i^{(u,v)}=J_{b})
-p(s_i^{(u,v)}=J_{a})\cdot p(\hat{y}_i^{(u,v)}=J_{b}),
\end{autobreak}
\end{align}
where $\forall a,b=\{1,2\}$ denote the entries of $\Delta$, $J_{1}=+1, J_{2}=-1$ represent the salient object and background class labels for pixel $(u,v)$ and $p(\cdot)$ represents the distribution. 
\end{myDef}

$\Delta_{a,b}$ in the Definition 1 captures the stochastic correlation between the pixel in the initial predicted saliency maps and noisy labels, which can be regarded as a loss criterion. Furthermore, we describe a binary scoring matrix $\Omega$ to indicate the specific correlation between pixels in the initial predicted saliency maps and noisy labels.

\begin{myDef}
The scoring matrix $\Omega$ is computed as:
\begin{equation}
\Omega(s_i^{(u,v)},\hat{y}_i^{(u,v)})= {\rm Sgn}(\Delta_{a,b}),
\end{equation}
where ${\rm Sgn}(\Delta_{a,b})=1$ when $\Delta_{a,b}>0$ and ${\rm Sgn}(\Delta_{a,b})=0$, otherwise.
\end{myDef}

CA requires each pixel in the predicted map to perform multiple tasks: compute the correlation with its  corresponding noisy label and exploit the correlation between predictions of other scenes and unpaired noisy labels as the penalty to current scene. Ultimately the scoring function for each task, is defined as follows: 
\begin{equation}
S(s_i^{(u,v)},\hat{y}_i^{(u,v)})=\Omega(s_i^{(u,v)},\hat{y}_i^{(u,v)})-\Omega(s_{i_1}^{(u,v)},\hat{y}_{i_2}^{(u,v)}),
\end{equation}

For each sample $(s_i^{(u,v)},\hat{y}_i^{(u,v)})$, randomly draw another two samples $(s_{i_1}^{(u,v)},\hat{y}_{i_1}^{(u,v)})$, $(s_{i_2}^{(u,v)},\hat{y}_{i_2}^{(u,v)})$ such that $i_1\neq i_2$. We will name $(s_{i_1}^{(u,v)},\hat{y}_{i_1}^{(u,v)})$ and $(s_{i_2}^{(u,v)},\hat{y}_{i_2}^{(u,v)})$ as $i$'s correlation samples. After pairing $s_{i_1}^{(u,v)}$ with $\hat{y}_{i_2}^{(u,v)}$ (two independent scenes $i_1$ and $i_2$), we define the scoring function $S(\cdot)$ for each sampled scene $s_i^{(u,v)}$.
The first term in the scoring function above evaluates the saliency prediction $s_i^{(u,v)}$ using noisy labels $\hat{y}_{i}^{(u,v)}$, the second term defined on two independent scenes $i_1$, $i_2$ punishes the predictor from overly agreeing with the noisy labels, which is a penalty score of the current pixel.

We know that noises in each label is asymmetric. Hence, we use a new scoring function to adjust the degree of penalty:
\begin{equation}
\Psi(s_i^{(u,v)},\hat{y}_i^{(u,v)})=\Omega(s_i^{(u,v)},\hat{y}_i^{(u,v)})-\alpha \Omega(s_{i_1}^{(u,v)},\hat{y}_{i_2}^{(u,v)}),
\end{equation}

Moreover, following~\cite{peerloss}, we compute the correlation between saliency maps and noisy labels by cross entropy loss to replace the scoring function $\Omega$ in Eq.(\ref{cross_entropy}), due to its adaptability in salient object detection.

According to the characteristics of pixel-level tasks, the number of pixels in each light field image is huge, so the saliency prediction result needs more detailed evaluations. However, evaluating the random correlation based only on a pair of cross-scene samples will cause large variance and is not stable enough. Therefore, we eliminate the variance as much as possible based on $m_l$ pairs of cross-scene samples to stabilize the training process:
\vspace{-2mm}
\begin{align}
\begin{autobreak}
\mathcal{L}_t(s_i^{(u,v)},\hat{y}_i^{(u,v)})=
\mathcal{L}(s_i^{(u,v)},\hat{y}_i^{(u,v)})
-\frac{\alpha}{m_l-1}\sum_{n,n'=2}^{m_l}(l(s_{i_n}^{(u,v)},\hat{y}_{i_{n'}}^{(u,v)})),
\end{autobreak}
\end{align}

Where we set $\alpha=0.2$ and $m_l=4$ in our experiment. The first term of our cross-scene noise penalty loss $\mathcal{L}_t$ evaluates the network's prediction on training data using noisy labels, and the second term is defined on several independent randomly selected light field images to penalize the networks from overly agreeing with the pixel-level noisy labels. Both terms encode the knowledge of noise rates implicitly and allow our light field  saliency prediction model to become invariant to pixel-level noise.

\begin{table*}[]
\centering
\resizebox{\textwidth}{!}{%
\begin{tabular}{c|c|cccccccccc|cc|cccc|cc}
\hline
\multirow{4}{*}{Dataset} & \multirow{4}{*}{Metrics} & \multicolumn{10}{c|}{Fully supervised Models} & \multicolumn{2}{c|}{Conventional Models} & \multicolumn{4}{c|}{Multi noisy labels Models} & \multicolumn{2}{c}{Single noisy label Models} \\ \cline{3-20} 
 &  & \multicolumn{4}{c|}{RGB} & \multicolumn{4}{c|}{RGB-D} & \multicolumn{2}{c|}{Light field} & \multicolumn{1}{c|}{RGB} & Light field & \multicolumn{4}{c|}{RGB} & \multicolumn{1}{c|}{RGB} & Light filed \\
 &  & PoolNet & PiCANet & R3Net & \multicolumn{1}{c|}{NLDF} & TANet & PCA & MMCI & \multicolumn{1}{c|}{UCNet} & DLLF & Mo-LF & \multicolumn{1}{c|}{RBD} & LFS & SBF & DUSPS & MNL & NAED & \multicolumn{1}{c|}{SNL} & Ours  \\
 &  & \cite{PoolNet} & \cite{PiCANet} & \cite{R3Net} & \multicolumn{1}{c|}{\cite{NLDF}} & \cite{TANet} & \cite{PCA} & \cite{MMCI} & \multicolumn{1}{c|}{\cite{UCNet}} & \cite{DLLF} & \cite{Memory_Nips} & \multicolumn{1}{c|}{\cite{RBD}} & \cite{LFS} & \cite{RGB_salience_fusion} & \cite{RGB_salience_nips} & \cite{RGB_salience_multi} & \cite{RGB_salience_Eccv} & \multicolumn{1}{c|}{\cite{RGB_salience_PAMI}} &  \\ \hline
\multirow{2}{*}{DUT-LF} & F$\uparrow$ & 0.868 & 0.821 & 0.783 & \multicolumn{1}{c|}{0.778} & 0.771 & 0.762 & 0.750 & \multicolumn{1}{c|}{0.819} & 0.868 & 0.843 & \multicolumn{1}{c|}{0.631} & 0.484 & 0.583 & 0.736 & 0.716 & 0.701 & \multicolumn{1}{c|}{0.679} & 0.813 \\
 & M$\downarrow$ & 0.051 & 0.083 & 0.113 & \multicolumn{1}{c|}{0.103} & 0.096 & 0.100 & 0.116 & \multicolumn{1}{c|}{0.087} & 0.070 & 0.052 & \multicolumn{1}{c|}{0.212} & 0.240 & 0.135 & 0.062 & 0.086 & 0.070 & \multicolumn{1}{c|}{0.072} & 0.091 \\ \hline
\multirow{2}{*}{HFUT} & F$\uparrow$ & 0.683 & 0.618 & 0.625 & \multicolumn{1}{c|}{0.636} & 0.605 & 0.619 & 0.645 & \multicolumn{1}{c|}{0.724} & 0.863 & 0.627 & \multicolumn{1}{c|}{0.601} & 0.430 & - & 0.705 & - & - & \multicolumn{1}{c|}{0.633} & 0.652 \\
 & M$\downarrow$ & 0.092 & 0.115 & 0.151 & \multicolumn{1}{c|}{0.091} & 0.111 & 0.104 & 0.104 & \multicolumn{1}{c|}{0.105} & 0.093 & 0.095 & \multicolumn{1}{c|}{0.241} & 0.205 & - & 0.087 & - & - & \multicolumn{1}{c|}{0.165} & 0.108 \\ \hline
\multirow{2}{*}{LFSD} & F$\uparrow$ & 0.769 & 0.671 & 0.781 & \multicolumn{1}{c|}{0.748} & 0.804 & 0.801 & 0.796 & \multicolumn{1}{c|}{0.835} & - & 0.819 & \multicolumn{1}{c|}{0.711} & 0.715 & - & 0.795 & - & - & \multicolumn{1}{c|}{0.714} & 0.804 \\
 & M$\downarrow$ & 0.118 & 0.158 & 0.128 & \multicolumn{1}{c|}{0.138} & 0.112 & 0.112 & 0.128 & \multicolumn{1}{c|}{0.108} & - & 0.089 & \multicolumn{1}{c|}{0.182} & 0.147 & - & 0.105 & - & - & \multicolumn{1}{c|}{0.097} & 0.111 \\ \hline
\end{tabular}%
}
\vspace{-2mm}
\caption{Quantitative comparisons on three light field datasets. $\uparrow$ $\&$ $\downarrow$ denote larger and smaller is better respectively}
\label{Table_compare}
\vspace{-3mm}
\end{table*}

\vspace{-2mm}
\section{Experimental Results}
\vspace{-2mm}

Our experiment are conducted on three public light field benchmark datasets: DUT-LF\cite{DUT-LF}, HFUT\cite{HFUT} and LFSD\cite{LFSD}. DUT-LF is proposed for fully supervised saliency detection containing 1462 challenging scenes with high similarity between the salient object and background, small-scale salient objects and various lighting conditions. DUT-LF is split into 1000 training and 462 test samples. HFUT and LFSD are relatively small, containing only 255 and 100 samples respectively. HFUT, LFSD and the test partition of DUT-LF are used to evaluate the performance of our method. All three datasets are captured with the Lytro camera~\cite{Lytro} and include an all-focus central view image, a stack of focal slices and the corresponding pixel-level annotated ground truth (GT) saliency map. Note that GT maps are depicted only for illustration purposes in this paper and not used in our training process.

\vspace{-1mm}
\subsection{Implementation Details}
\vspace{-1mm}
Our model is implemented in PyTorch and trained for a maximum of 30 epochs using a single GeForce GTX TITAN X GPU. We train the model in an end-to-end manner using 0.9 momentum and a learning rate of $1.0\times10^{-5}$. We use the Adam\cite{Adam} optimizer with the 'Inverse' decay policy. We initialize the RGB and focal stack streams using VGG-19~\cite{VGG} trained for image classification, and adapt it to our task. We also augment the training data with random flipping, cropping and rotation. For the handcrafted methods, we generate pixel-level noisy saliency maps using the RBD method~\cite{RBD}, due to its high efficiency. Similar to prior works, we use F-measure~\cite{F-measure} and Mean Absolute Error (MAE) as the evaluation metrics for comprehensive bench-marking of our algorithm.

\begin{figure*}
	\centering
	\includegraphics[width=\linewidth]{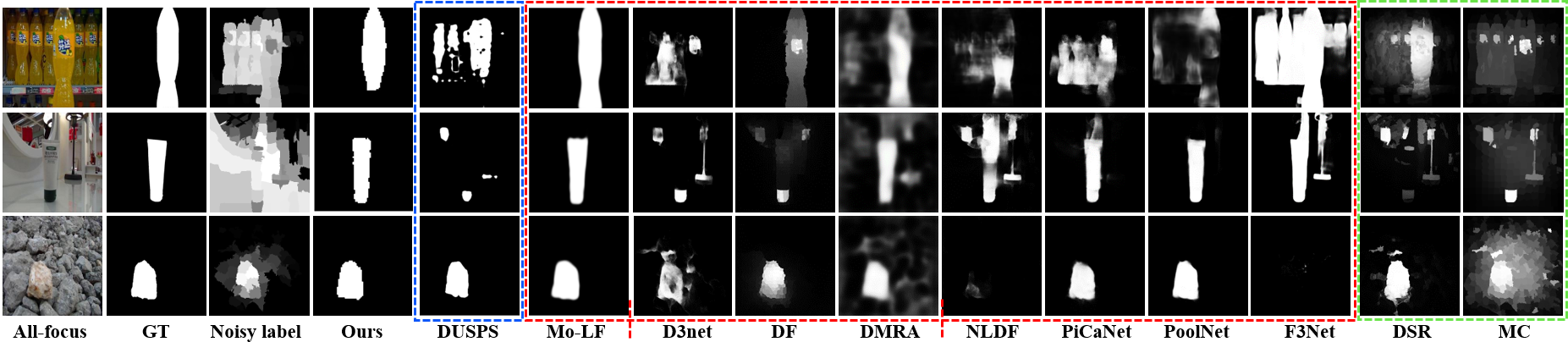}
	\vspace{-6mm}
	\caption{Visual comparison of the saliency maps with competing methods, where the saliency maps in the blue box are predicted from noisy labels supervised RGB methods, the saliency maps in the red box are predicted from fully supervised light field, RGB-D and RGB saliency method respectively and the saliency maps in the green box are predicted from conventional models.}
	\label{Fig_vis}
	\vspace{-3mm}
\end{figure*}

\begin{table}
\small
\centering
\fontsize{8}{10}\selectfont
\setlength{\tabcolsep}{3mm}{
\begin{tabular}{c|cccc}
\toprule
Settings & Metrics & DUT-LF & HUFT & LFSD \\ \hline
\multirow{2}{*}{Baseline}&$F\uparrow$ & 0.641 & 0.553 & 0.683 \\
 & $M\downarrow$ & 0.279 & 0.253 & 0.191 \\ \hline
 \multirow{2}{*}{+ MFFO} & $F\uparrow$ & 0.689 & 0.611 & 0.737 \\
 & $M\downarrow$ & 0.214 & 0.195 & 0.165 \\ \hline
\multirow{2}{*}{+ PFM} & $F\uparrow$ & 0.741 & 0.634 & 0.760 \\
 & $M\downarrow$ & 0.181 & 0.163 & 0.179 \\ \hline
\multirow{2}{*}{+ Ploss} & $F\uparrow$ & 0.730 & 0.620 & 0.749 \\ 
& $M\downarrow$ & 0.147 & 0.151 & 0.132 \\ \hline
\multirow{2}{*}{\textbf{Ours}} & $F\uparrow$ & \textbf{0.813} & \textbf{0.652} & \textbf{0.804} \\
 & $M\downarrow$ & \textbf{0.091} & \textbf{0.108} & \textbf{0.111} \\ 
\bottomrule
\end{tabular}}
\vspace{-2mm}
\caption{Results on extensive ablation studies analyzing the significance of different components on our pipeline.}
\label{Table_ablation}
\vspace{-3mm}
\end{table}

\begin{table}
  \centering
  \fontsize{7.5}{10}\selectfont
    \begin{tabular}{c|cccccccc}
    \toprule
    \multirow{1}{*}{$\delta$}
    &0.1&0.2&{\bf 0.3}&0.4&0.5&0.6&0.7\\
    \midrule
    \multirow{1}*{$F\uparrow$} & 0.744  & 0.786 &{\bf 0.813} & 0.801 & 0.792 & 0.761 & 0.771 \\
    \multirow{1}{*}{$M\downarrow$} & 0.221 & 0.183 & {\bf0.091} & 0.091 & 0.163 & 0.155 & 0.160 \\            
    \bottomrule
    \end{tabular}
    \vspace{-2mm}
    \caption{Comparing the affect of varying number of $\delta$ in Eq.(\ref{Euqation_delta}) on saliency detection performance keeping the other settings of our framework unchanged.}
   \label{Table_delta}
   \vspace{-4mm}
\end{table}

\begin{figure}
	\centering
	\includegraphics[width=0.97\linewidth]{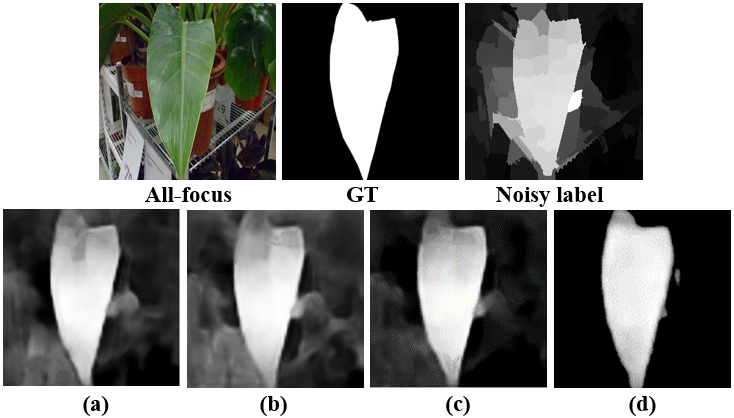}
	\vspace{-2mm}
	\caption{Visualization of the saliency maps at different steps.(a)-(b) represent the results from the refined focal slices and all focus features respectively, (c)-(d) represent the results from pixel forgetting guided features fusion and cross-scene noise penalty loss.}
	\label{Fig_vis_steps}
	\vspace{-4mm}
\end{figure}

\begin{figure*}
	\centering
	\includegraphics[width=\linewidth]{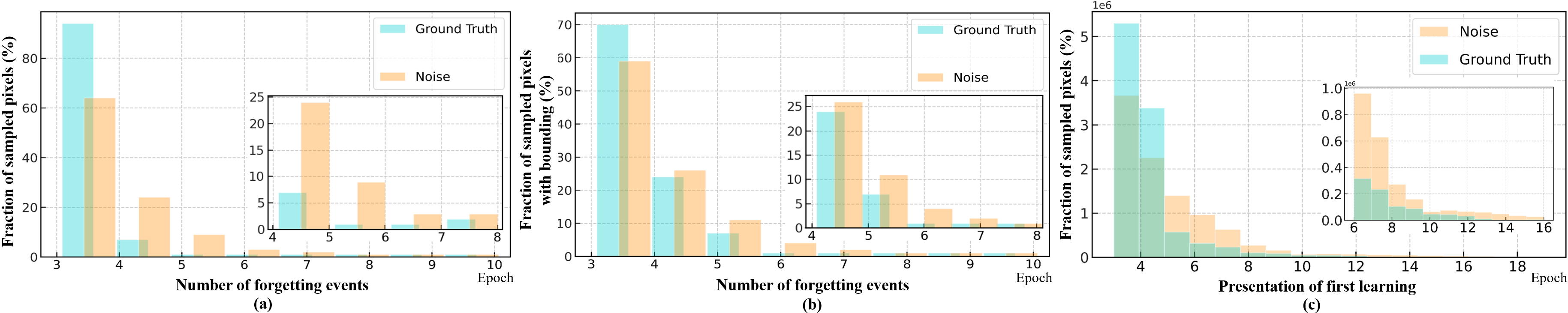}
	\vspace{-6mm}
	\caption{
	The distributions of forgetting events and first learning.
	}
	\label{Fig_forg_three}
	\vspace{-4mm}
\end{figure*}

\vspace{-1mm}
\subsection{Comparison with Start-of-the-art Methods}
\vspace{-2mm}
For a comprehensive evaluation, we compare our method with 23 state-of-the-art saliency detection models, including 5 fully supervised RGB methods (PoolNet~\cite{PoolNet}, PiCANet~\cite{PiCANet}, R3Net~\cite{R3Net}, NLDF~\cite{NLDF}, F3Net~\cite{F3Net}), 7 supervised RGB-D methods (TANet~\cite{TANet}, PCA~\cite{PCA}, MMCI~\cite{MMCI}, UCNet~\cite{UCNet}, DF~\cite{DF}, DMRA~\cite{DMRA}, D3Net~\cite{D3Net}), 
2 supervised light field methods (DLLF~\cite{DLLF}, Mo-LF~\cite{Memory_Nips}), 4 conventional unsupervised methods (RBD~\cite{RBD}, LFS~\cite{LFS},DSR~\cite{DSR}, MC~\cite{MC}),
4 multi noisy labels supervised methods (SBF~\cite{RGB_salience_fusion}, DUSPS~\cite{RGB_salience_nips}, MNL~\cite{RGB_salience_multi}, NAED~\cite{RGB_salience_Eccv}) and 1 single noisy label supervised method
SNL~\cite{RGB_salience_PAMI}.
%
Results of competing methods are generated by authorized codes or directly provided by authors.

\noindent\textbf{Quantitative Results} are shown in Table~\ref{Table_compare}. Compared with most fully supervised RGB saliency detection methods, our model consistently achieves higher scores on all datasets across two evaluation metrics. One important observation should be noted: although our model is supervised by pixel-level noisy labels, it still achieves significant advantage indicating that light field data are significant and promising due to its abundant spatial information. The effectiveness of light field data is further supported by the superior improvement compared with some fully supervised RGB-D methods. Compared with the coarse depth maps in RGB-D data, light field contains more accurate depth cues. More encouragingly, when compared with a number of fully supervised light field methods, our method still achieves competitive performance. These performances are reasonable since the effective forgetting pixel guided features fusion and the proper handling of correlations across scenes.

Moreover, we can see that our method outperforms the conventional RGB method RBD~\cite{RBD} with a significant margin of 0.18 in the F-measure on the DUT-LF dataset, which is used to generate the noisy labels for our model. This is mainly because our method explore the intra light field features fusion and inter scenes correlation to generate the robust predictions. We also compare our method with the state-of-the-art noisy label supervised RGB models, we can see that our method leads to an improvement of up to 0.23 in the F-measure metric on DUT-LF dataset, demonstrating the superiority of the abundant light field cues, as well as our proposed features fusion and noise penalty strategies.

\noindent\textbf{Qualitative Comparison} is given in Figure~\ref{Fig_vis} where we visualize three representative saliency map comparison cases. We see that our method is able to handle a wide range of challenging scenes, including similar foreground and background (first row), clutter background (second row) and small object (third row). Our method can predict the salient objects with relatively complete boundary information even the provided noisy labels are incomplete in salient objects, which is an exciting breakthrough. Compared with the noisy label supervised RGB saliency method DUSPS~\cite{RGB_salience_nips}, our model not only more accurately localizes the salient objects, but also more precisely recovers the object details, which are positively influenced by the light field data and our proposed modules. Our method also achieves competitive detection results compared with the fully supervised light field method Mo-LF~\cite{Memory_Nips}. More qualitative comparisons can be found in the supplementary material.

\begin{figure}
	\centering
	\includegraphics[width=0.8\linewidth]{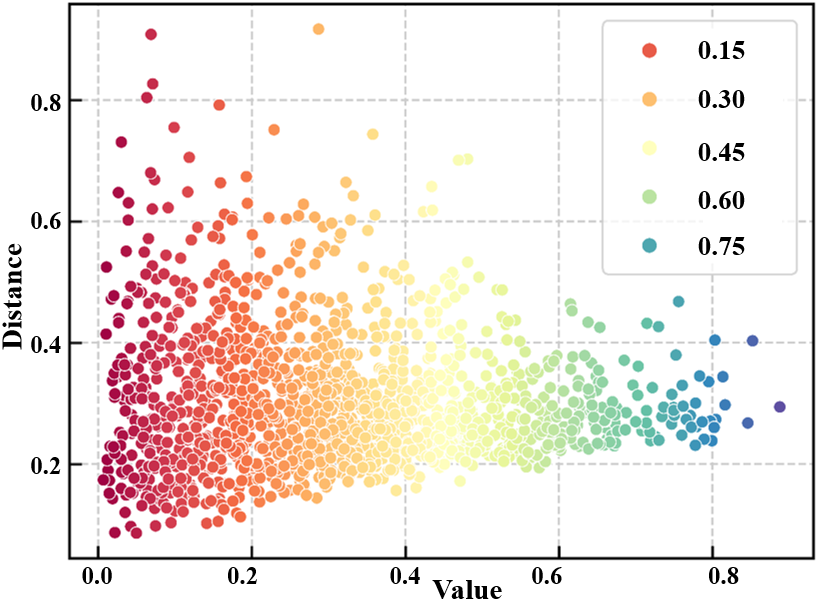}
	\vspace{-3mm}
	\caption{Characteristic correlations among noisy pixels in labels across the DUT-LF dataset~\cite{DUT-LF}.
	}
	\label{Fig_relation}
	\vspace{-6mm}
\end{figure}

\vspace{-2mm}
\subsection{Ablation Studies}
\vspace{-2mm}
We conduct ablation experiments on DUT-LF~\cite{DUT-LF} dataset to thoroughly analyze the effectiveness of our proposed modules. To simplify the experiments and get more intuitive results, we build a concise baseline that only contains separate features extraction branches for focal slices and all-focus central view image. The saliency map is predicted by directly concatenating the two kinds of features.

\noindent\textbf{1) Multual Feature Fusion Operation (MFFO):} As shown in Table~\ref{Table_ablation}, the proposed mutual feature fusion operation results in improved performance which means that the interactions among all-focus central view image and focal slices explore the abundant spatial information of light field data, which are necessary to predict the salient object. 

\noindent\textbf{2) Pixel Forgetting Matrix (PFM):}
To evaluate the pixel consistency of the initial predicted noisy saliency maps from all-focus central view image and focal slices during training iterations, we propose the pixel forgetting matrix and analyse whether there exist noisy pixels that are consistently forgotten across subsequent training presentation and, conversely, pixels that are rarely forgotten. From Table ~\ref{Table_ablation} we can see that the pixel forgetting matrix improves the accuracy of saliency detection. Moreover, we define $\delta$ as a threshold to identify the pixel is learned or forgotten. To explore the optimal threshold, we conduct experiments with $\delta$ in a range of (0,1) and report performances in Table~\ref{Table_delta}. Our evaluation achieves highest scores when $\delta=0.3$.

\noindent\textbf{3) Cross-scene Noise Penalty loss (Ploss)}.
As shown in Table~\ref{Table_ablation}, we demonstrate the effect of the proposed cross-scene noise penalty loss. Compared with the baseline, we observe that it can boost our performance by a large margin (30\% improvement in MAE metric), which further illustrates that the loss effectively penalizes the networks from overly agreeing with the pixel-level noisy labels.

\noindent\textbf{4) Visualization of Intermediate Results}.
In Figure~\ref{Fig_vis_steps}, we show an example saliency detection result to illustrate the performances with respect to our proposed modules. Starting with the noisy label supervision, our method consistently improves the performance of saliency detection with the cumulative updating of proposed modules.

\vspace{-1mm}
\subsection{Further Analysis}
\vspace{-2mm}
\noindent\textbf{Pixel forgetting:} To validate our assumption that noisy pixels experience frequently forgetting events compared with consistent pixels, we train our model once using pixel-level noisy labels and again using ground truth labels. The distribution of forgetting events across the fraction of the noisy pixels and consistent pixels are shown in Fig.~\ref{Fig_forg_three}(a). Specifically, we sample the pixels from the dataset and describe the relative fraction of pixels which experience over 3 forgetting events during training. We further sample pixels in the coarse bounding box of the salient object 
%
%
and report results in Fig.~\ref{Fig_forg_three}(b). We observe a relatively higher degree of forgetting in noisy supervision with more obvious forgetting of noise within the salient object region.

We also investigate the occurrence time of first learning events of noisy labels and ground truth. The distribution of average occurrence time at which the first learning events occur across sampled pixels (over 5 seeds) of noisy labels and ground truth are shown in Fig.~\ref{Fig_forg_three}(c). 
Notice that most pixels of noisy labels and ground truth are both learned during the first 4 epochs, while the noisy pixels contain a larger number of pixels 
learned during later part of the training. 
%
Noisy pixels exhibit characteristics different from consistent pixels during training, which is important for our model to identify noise.

\noindent\textbf{Across-scene Latent Noise correlation:} 
We investigate whether noisy label characteristics are distributed regularly cross different scenes. Each point in Fig.~\ref{Fig_relation} represents one noisy label in the dataset i.e., average pixel value versus average distance. 
The intensity values of noisy pixels and their distance from the center of salient object are characteristics implicitly related in different scenes. The effectiveness of our cross scene penalty loss is reasonable given the latent noise correlation in the noisy dataset.

\vspace{-3mm}
\section{Conclusion}
\vspace{-2mm}
In this paper, we represent, for the first time, light field saliency detection as a learning from pixel-level noisy labels problem. This allows us to use efficient conventional unsupervised saliency detection methods. To leverage the relationships among light field cues while identifying clean labels in a unified framework, we proposed pixel forgetting guided fusion to mutually enhance the light field features and exploit pixel consistency across iterations to identify noisy pixel labels. We also proposed cross scene noise penalty loss to better reflect latent structures of training data enabling the learning to be invariant to noise. Extended experiments show the superiority of our method, which not only outperforms most RGB and RGB-D saliency methods but also achieves comparable performance to state-of-the-art fully supervised light field saliency detection methods.

\vspace{-3mm}
\section{Acknowledgments}
\vspace{-2mm}
Professor Ajmal Mian is the recipient of an Australian Research Council Future Fellowship Award (project number FT210100268) funded by the Australian Government.









\normalem
{\small
\bibliographystyle{ieee_fullname}
\bibliography{egbib}

\begin{thebibliography}{10}\itemsep=-1pt

\bibitem{Lytro}
http://lightfield-forum.com/lytro/lytro-illum-professional-light-field-camera/.

\bibitem{Raytrix}
https://raytrix.de/.

\bibitem{F-measure}
Radhakrishna Achanta, Sheila Hemami, Francisco Estrada, and Sabine Susstrunk.
\newblock Frequency-tuned salient region detection.
\newblock In {\em 2009 IEEE conference on computer vision and pattern
  recognition}, pages 1597--1604. IEEE, 2009.

\bibitem{PCA}
Hao Chen and Youfu Li.
\newblock Progressively complementarity-aware fusion network for rgb-d salient
  object detection.
\newblock In {\em Proceedings of the IEEE conference on computer vision and
  pattern recognition}, pages 3051--3060, 2018.

\bibitem{TANet}
Hao Chen and Youfu Li.
\newblock Three-stream attention-aware network for rgb-d salient object
  detection.
\newblock {\em IEEE Transactions on Image Processing}, 28(6):2825--2835, 2019.

\bibitem{MMCI}
Hao Chen, Youfu Li, and Dan Su.
\newblock Multi-modal fusion network with multi-scale multi-path and
  cross-modal interactions for rgb-d salient object detection.
\newblock {\em Pattern Recognition}, 86:376--385, 2019.

\bibitem{SOD_review1}
Runmin Cong, Jianjun Lei, Huazhu Fu, Ming-Ming Cheng, Weisi Lin, and Qingming
  Huang.
\newblock Review of visual saliency detection with comprehensive information.
\newblock {\em IEEE Transactions on circuits and Systems for Video Technology},
  29(10):2941--2959, 2018.

\bibitem{CA2}
Anirban Dasgupta and Arpita Ghosh.
\newblock Crowdsourced judgement elicitation with endogenous proficiency.
\newblock In {\em Proceedings of the 22nd international conference on World
  Wide Web}, pages 319--330, 2013.

\bibitem{R3Net}
Zijun Deng, Xiaowei Hu, Lei Zhu, Xuemiao Xu, Jing Qin, Guoqiang Han, and
  Pheng-Ann Heng.
\newblock R3net: Recurrent residual refinement network for saliency detection.
\newblock In {\em Proceedings of the 27th International Joint Conference on
  Artificial Intelligence}, pages 684--690, 2018.

\bibitem{D3Net}
Deng-Ping Fan, Zheng Lin, Zhao Zhang, Menglong Zhu, and Ming-Ming Cheng.
\newblock Rethinking rgb-d salient object detection: Models, data sets, and
  large-scale benchmarks.
\newblock {\em IEEE Transactions on neural networks and learning systems},
  32(5):2075--2089, 2020.

\bibitem{LF_1}
Mingtao Feng, Yaonan Wang, Jian Liu, Liang Zhang, Hasan~FM Zaki, and Ajmal
  Mian.
\newblock Benchmark data set and method for depth estimation from light field
  images.
\newblock {\em IEEE Transactions on Image Processing}, 27(7):3586--3598, 2018.

\bibitem{LF_review}
Keren Fu, Yao Jiang, Ge-Peng Ji, Tao Zhou, Qijun Zhao, and Deng-Ping Fan.
\newblock Light field salient object detection: A review and benchmark.
\newblock {\em arXiv preprint arXiv:2010.04968}, 2020.

\bibitem{SOD_review2}
Stas Goferman, Lihi Zelnik-Manor, and Ayellet Tal.
\newblock Context-aware saliency detection.
\newblock {\em IEEE transactions on pattern analysis and machine intelligence},
  34(10):1915--1926, 2011.

\bibitem{Noisy_2}
Jacob Goldberger and Ehud Ben-Reuven.
\newblock Training deep neural-networks using a noise adaptation layer.
\newblock In {\em International Conference on Learning Representations}, 2016.

\bibitem{MC}
Bowen Jiang, Lihe Zhang, Huchuan Lu, Chuan Yang, and Ming-Hsuan Yang.
\newblock Saliency detection via absorbing markov chain.
\newblock In {\em Proceedings of the IEEE international conference on computer
  vision}, pages 1665--1672, 2013.

\bibitem{Noisy_4}
Lu Jiang, Zhengyuan Zhou, Thomas Leung, Li-Jia Li, and Li Fei-Fei.
\newblock Mentornet: Learning data-driven curriculum for very deep neural
  networks on corrupted labels.
\newblock In {\em International Conference on Machine Learning}, pages
  2304--2313, 2018.

\bibitem{Adam}
Diederik~P Kingma and Jimmy Ba.
\newblock Adam: A method for stochastic optimization.
\newblock {\em arXiv preprint arXiv:1412.6980}, 2014.

\bibitem{T_LFSaliency_2014}
Nianyi Li, Jinwei Ye, Yu Ji, Haibin Ling, and Jingyi Yu.
\newblock Saliency detection on light field.
\newblock In {\em Proceedings of the IEEE Conference on Computer Vision and
  Pattern Recognition}, pages 2806--2813, 2014.

\bibitem{LFS}
Nianyi Li, Jinwei Ye, Yu Ji, Haibin Ling, and Jingyi Yu.
\newblock Saliency detection on light field.
\newblock In {\em Proceedings of the IEEE Conference on Computer Vision and
  Pattern Recognition}, pages 2806--2813, 2014.

\bibitem{LFSD}
Nianyi Li, Jinwei Ye, Yu Ji, Haibin Ling, and Jingyi Yu.
\newblock Saliency detection on light field.
\newblock In {\em Proceedings of the IEEE Conference on Computer Vision and
  Pattern Recognition}, pages 2806--2813, 2014.

\bibitem{DSR}
Xiaohui Li, Huchuan Lu, Lihe Zhang, Xiang Ruan, and Ming-Hsuan Yang.
\newblock Saliency detection via dense and sparse reconstruction.
\newblock In {\em Proceedings of the IEEE international conference on computer
  vision}, pages 2976--2983, 2013.

\bibitem{PoolNet}
Jiang-Jiang Liu, Qibin Hou, Ming-Ming Cheng, Jiashi Feng, and Jianmin Jiang.
\newblock A simple pooling-based design for real-time salient object detection.
\newblock In {\em Proceedings of the IEEE/CVF Conference on Computer Vision and
  Pattern Recognition}, pages 3917--3926, 2019.

\bibitem{PiCANet}
Nian Liu, Junwei Han, and Ming-Hsuan Yang.
\newblock Picanet: Learning pixel-wise contextual attention for saliency
  detection.
\newblock In {\em Proceedings of the IEEE Conference on Computer Vision and
  Pattern Recognition}, pages 3089--3098, 2018.

\bibitem{LFSOD_CVPR21}
Nian Liu, Wangbo Zhao, Dingwen Zhang, Junwei Han, and Ling Shao.
\newblock Light field saliency detection with dual local graph learning and
  reciprocative guidance.
\newblock In {\em Proceedings of the IEEE/CVF International Conference on
  Computer Vision}, pages 4712--4721, 2021.

\bibitem{Noisy_5}
Tongliang Liu and Dacheng Tao.
\newblock Classification with noisy labels by importance reweighting.
\newblock {\em IEEE Transactions on pattern analysis and machine intelligence},
  38(3):447--461, 2015.

\bibitem{peerloss}
Yang Liu and Hongyi Guo.
\newblock Peer loss functions: Learning from noisy labels without knowing noise
  rates.
\newblock In {\em International Conference on Machine Learning}, pages
  6226--6236. PMLR, 2020.

\bibitem{NLDF}
Zhiming Luo, Akshaya Mishra, Andrew Achkar, Justin Eichel, Shaozi Li, and
  Pierre-Marc Jodoin.
\newblock Non-local deep features for salient object detection.
\newblock In {\em Proceedings of the IEEE Conference on computer vision and
  pattern recognition}, pages 6609--6617, 2017.

\bibitem{RGB_salience_nips}
Duc~Tam Nguyen, Maximilian Dax, Chaithanya~Kumar Mummadi, Thi Phuong~Nhung Ngo,
  Thi Hoai~Phuong Nguyen, Zhongyu Lou, and Thomas Brox.
\newblock Deepusps: deep robust unsupervised saliency prediction with
  self-supervision.
\newblock In {\em Proceedings of the 33rd International Conference on Neural
  Information Processing Systems}, pages 204--214, 2019.

\bibitem{Noisy_7}
Tam Nguyen, C Mummadi, T Ngo, L Beggel, and Thomas Brox.
\newblock Self: learning to filter noisy labels with self-ensembling.
\newblock In {\em International Conference on Learning Representations (ICLR)},
  2020.

\bibitem{DMRA}
Yongri Piao, Wei Ji, Jingjing Li, Miao Zhang, and Huchuan Lu.
\newblock Depth-induced multi-scale recurrent attention network for saliency
  detection.
\newblock In {\em Proceedings of the IEEE/CVF International Conference on
  Computer Vision}, pages 7254--7263, 2019.

\bibitem{LFSOD_cellular}
Yongri Piao, Xiao Li, Miao Zhang, Jingyi Yu, and Huchuan Lu.
\newblock Saliency detection via depth-induced cellular automata on light
  field.
\newblock {\em IEEE Transactions on Image Processing}, 29:1879--1889, 2019.

\bibitem{T_LFSaliency_TIP}
Yongri Piao, Xiao Li, Miao Zhang, Jingyi Yu, and Huchuan Lu.
\newblock Saliency detection via depth-induced cellular automata on light
  field.
\newblock {\em IEEE Transactions on Image Processing}, 29:1879--1889, 2019.

\bibitem{Exploit_LF_fusion}
Yongri Piao, Zhengkun Rong, Miao Zhang, and Huchuan Lu.
\newblock Exploit and replace: An asymmetrical two-stream architecture for
  versatile light field saliency detection.
\newblock In {\em Proceedings of the AAAI Conference on Artificial
  Intelligence}, volume~34, pages 11865--11873, 2020.

\bibitem{DF}
Liangqiong Qu, Shengfeng He, Jiawei Zhang, Jiandong Tian, Yandong Tang, and
  Qingxiong Yang.
\newblock Rgbd salient object detection via deep fusion.
\newblock {\em IEEE Transactions on Image Processing}, 26(5):2274--2285, 2017.

\bibitem{Noisy_6}
Mengye Ren, Wenyuan Zeng, Bin Yang, and Raquel Urtasun.
\newblock Learning to reweight examples for robust deep learning.
\newblock In {\em International Conference on Machine Learning}, pages
  4334--4343, 2018.

\bibitem{CA1}
Victor Shnayder, Arpit Agarwal, Rafael Frongillo, and David~C Parkes.
\newblock Informed truthfulness in multi-task peer prediction.
\newblock In {\em Proceedings of the 2016 ACM Conference on Economics and
  Computation}, pages 179--196, 2016.

\bibitem{VGG}
Karen Simonyan and Andrew Zisserman.
\newblock Very deep convolutional networks for large-scale image recognition.
\newblock {\em arXiv preprint arXiv:1409.1556}, 2014.

\bibitem{Noisy_3}
Ryutaro Tanno, Ardavan Saeedi, Swami Sankaranarayanan, Daniel~C Alexander, and
  Nathan Silberman.
\newblock Learning from noisy labels by regularized estimation of annotator
  confusion.
\newblock In {\em Proceedings of the IEEE/CVF Conference on Computer Vision and
  Pattern Recognition}, pages 11244--11253, 2019.

\bibitem{Forgetting}
Mariya Toneva, Alessandro Sordoni, Remi~Tachet des Combes, Adam Trischler,
  Yoshua Bengio, and Geoffrey~J Gordon.
\newblock An empirical study of example forgetting during deep neural network
  learning.
\newblock In {\em International Conference on Learning Representations}, 2018.

\bibitem{Deep_LF_fusion}
Tiantian Wang, Yongri Piao, Xiao Li, Lihe Zhang, and Huchuan Lu.
\newblock Deep learning for light field saliency detection.
\newblock In {\em Proceedings of the IEEE/CVF International Conference on
  Computer Vision}, pages 8838--8848, 2019.

\bibitem{DLLF}
Tiantian Wang, Yongri Piao, Xiao Li, Lihe Zhang, and Huchuan Lu.
\newblock Deep learning for light field saliency detection.
\newblock In {\em Proceedings of the IEEE/CVF International Conference on
  Computer Vision}, pages 8838--8848, 2019.

\bibitem{DUT-LF}
Tiantian Wang, Yongri Piao, Xiao Li, Lihe Zhang, and Huchuan Lu.
\newblock Deep learning for light field saliency detection.
\newblock In {\em Proceedings of the IEEE/CVF International Conference on
  Computer Vision}, pages 8838--8848, 2019.

\bibitem{PAMI_CRF}
Yang Wang, Kia-Fock Loe, and Jian-Kang Wu.
\newblock A dynamic conditional random field model for foreground and shadow
  segmentation.
\newblock {\em IEEE transactions on pattern analysis and machine intelligence},
  28(2):279--289, 2005.

\bibitem{F3Net}
Jun Wei, Shuhui Wang, and Qingming Huang.
\newblock F$^3$net: Fusion, feedback and focus for salient object detection.
\newblock In {\em Proceedings of the AAAI Conference on Artificial
  Intelligence}, volume~34, pages 12321--12328, 2020.

\bibitem{ConvLSTM}
SHI Xingjian, Zhourong Chen, Hao Wang, Dit-Yan Yeung, Wai-Kin Wong, and
  Wang-chun Woo.
\newblock Convolutional lstm network: A machine learning approach for
  precipitation nowcasting.
\newblock In {\em Advances in neural information processing systems}, pages
  802--810, 2015.

\bibitem{Noisy_1}
Kun Yi and Jianxin Wu.
\newblock Probabilistic end-to-end noise correction for learning with noisy
  labels.
\newblock In {\em Proceedings of the IEEE/CVF Conference on Computer Vision and
  Pattern Recognition}, pages 7017--7025, 2019.

\bibitem{Noise_network}
Chiyuan Zhang, Samy Bengio, Moritz Hardt, Benjamin Recht, and Oriol Vinyals.
\newblock Understanding deep learning (still) requires rethinking
  generalization.
\newblock {\em Communications of the ACM}, 64(3):107--115, 2021.

\bibitem{RGB_salience_fusion}
Dingwen Zhang, Junwei Han, and Yu Zhang.
\newblock Supervision by fusion: Towards unsupervised learning of deep salient
  object detector.
\newblock In {\em Proceedings of the IEEE International Conference on Computer
  Vision}, pages 4048--4056, 2017.

\bibitem{RGB_salience_PAMI}
Jing Zhang, Yuchao Dai, Tong Zhang, Mehrtash~T Harandi, Nick Barnes, and
  Richard Hartley.
\newblock Learning saliency from single noisy labelling: A robust model fitting
  perspective.
\newblock {\em IEEE Transactions on Pattern Analysis and Machine Intelligence},
  2021.

\bibitem{UCNet}
Jing Zhang, Deng-Ping Fan, Yuchao Dai, Saeed Anwar, Fatemeh~Sadat Saleh, Tong
  Zhang, and Nick Barnes.
\newblock Uc-net: Uncertainty inspired rgb-d saliency detection via conditional
  variational autoencoders.
\newblock In {\em Proceedings of the IEEE/CVF conference on computer vision and
  pattern recognition}, pages 8582--8591, 2020.

\bibitem{SOD_review3}
Jing Zhang, Deng-Ping Fan, Yuchao Dai, Xin Yu, Yiran Zhong, Nick Barnes, and
  Ling Shao.
\newblock Rgb-d saliency detection via cascaded mutual information
  minimization.
\newblock In {\em Proceedings of the IEEE/CVF International Conference on
  Computer Vision}, pages 4338--4347, 2021.

\bibitem{T_LFSaliency_Ijcai15}
Jun Zhang, Meng Wang, Jun Gao, Yi Wang, Xudong Zhang, and Xindong Wu.
\newblock Saliency detection with a deeper investigation of light field.
\newblock In {\em Twenty-Fourth International Joint Conference on Artificial
  Intelligence}, 2015.

\bibitem{HFUT}
Jun Zhang, Meng Wang, Liang Lin, Xun Yang, Jun Gao, and Yong Rui.
\newblock Saliency detection on light field: A multi-cue approach.
\newblock {\em ACM Transactions on Multimedia Computing, Communications, and
  Applications (TOMM)}, 13(3):1--22, 2017.

\bibitem{RGB_salience_Eccv}
Jing Zhang, Jianwen Xie, and Nick Barnes.
\newblock Learning noise-aware encoder-decoder from noisy labels by alternating
  back-propagation for saliency detection.
\newblock In {\em Computer Vision--ECCV 2020: 16th European Conference,
  Glasgow, UK, August 23--28, 2020, Proceedings, Part XVII 16}, pages 349--366,
  2020.

\bibitem{RGB_salience_multi}
Jing Zhang, Tong Zhang, Yuchao Dai, Mehrtash Harandi, and Richard Hartley.
\newblock Deep unsupervised saliency detection: A multiple noisy labeling
  perspective.
\newblock In {\em Proceedings of the IEEE conference on computer vision and
  pattern recognition}, pages 9029--9038, 2018.

\bibitem{LFSOD_LFNet}
Miao Zhang, Wei Ji, Yongri Piao, Jingjing Li, Yu Zhang, Shuang Xu, and Huchuan
  Lu.
\newblock Lfnet: Light field fusion network for salient object detection.
\newblock {\em IEEE Transactions on Image Processing}, 29:6276--6287, 2020.

\bibitem{Memory_Nips}
Miao Zhang, Jingjing Li, Wei Ji, Yongri Piao, and Huchuan Lu.
\newblock Memory-oriented decoder for light field salient object detection.
\newblock In {\em Proceedings of the 33rd International Conference on Neural
  Information Processing Systems}, pages 898--908, 2019.

\bibitem{LFSOD_muti-task}
Qiudan Zhang, Shiqi Wang, Xu Wang, Zhenhao Sun, Sam Kwong, and Jianmin Jiang.
\newblock A multi-task collaborative network for light field salient object
  detection.
\newblock {\em IEEE Transactions on Circuits and Systems for Video Technology},
  31(5):1849--1861, 2020.

\bibitem{RBD}
Wangjiang Zhu, Shuang Liang, Yichen Wei, and Jian Sun.
\newblock Saliency optimization from robust background detection.
\newblock In {\em Proceedings of the IEEE conference on computer vision and
  pattern recognition}, pages 2814--2821, 2014.

\end{thebibliography}
}

\end{document}